\documentclass{article} 

\usepackage{amsmath,amsfonts,bm}

\def\eqref#1{equation~\ref{#1}}

\def\1{\bm{1}}

\DeclareMathAlphabet{\mathsfit}{\encodingdefault}{\sfdefault}{m}{sl}
\SetMathAlphabet{\mathsfit}{bold}{\encodingdefault}{\sfdefault}{bx}{n}

\usepackage{hyperref}
\usepackage{url}
\usepackage{graphicx} 
\usepackage{caption,subcaption}

\usepackage[final]{iclr2022_conference}

\usepackage{natbib}
\setcitestyle{numbers,square,comma}

\usepackage[utf8]{inputenc} 
\usepackage[T1]{fontenc}    
\usepackage{hyperref}       
\usepackage{url}            
\usepackage{booktabs}       
\usepackage{amsfonts}       
\usepackage{nicefrac}       
\usepackage{microtype}      
\usepackage{xcolor}         
\usepackage{xspace}
\usepackage{graphicx}
\usepackage{amssymb}
\usepackage{color}
\usepackage{amsmath}

\usepackage{multirow}
\usepackage{wrapfig}
\usepackage[normalem]{ulem}
\usepackage{float}

\title{System III: Learning with Domain Knowledge for Safety Constraints}

\author{%
Fazl Barez$^{1}$ \quad Hosien Hasanbieg$^{2}$ \quad Alesandro Abbate$^2$\\
$^1$Edinburgh Centre for Robotics \quad $^2$University of Oxford \\

\texttt{\{f.barez@ed.ac.uk\}}\\
\texttt{\{hosein.hasanbeig, aabate\}@cs.ox.ac.uk}
}

\begin{document}

\maketitle

\begin{abstract}
Reinforcement learning agents naturally learn from extensive exploration. Exploration is costly and can be unsafe in \textit{safety-critical} domains. This paper proposes a novel framework for incorporating domain knowledge to help guide safe exploration and boost sample efficiency. Previous approaches impose constraints, such as regularisation parameters in neural networks, that rely on large sample sets and often are not suitable for safety-critical domains where agents should almost always avoid unsafe actions. In our approach, called \textit{System III}, which is inspired by psychologists' notions of the brain's \textit{System I} and \textit{System II},  
we represent domain expert knowledge of safety in form of first-order logic. We evaluate the satisfaction of these constraints via p-norms in state vector space. In our formulation, constraints are analogous to hazards, objects, and regions of state that have to be avoided during exploration.
We evaluated the effectiveness of the proposed method on OpenAI's Gym and Safety-Gym environments.
In all tasks, including classic Control and Safety Games, we show that our approach results in safer exploration and sample efficiency. 
\end{abstract}

\section{Introduction}

While existing Reinforcement Learning (RL) methods provide promising guarantees given sufficient exploration guarantees, in safety-critical applications most of the exploration methods are impractical due to the system vulnerability. Consider the example of a self-driving car where the controller agent should respect the speed limit, should not cross the stop sign, and should not crash into objects and other agents in the environment \citep{endtoend}. Safety in RL is not limited to self-driving cars, it can be used to make algorithms systematically safe and aligned with human intent \citep{concreateproblemsinmlsafety}. 
In \textit{safety-critical} domains such as autonomous driving, warehouse logistics or assistance in health care, experts require deep RL controllers to operate within known bounds and limits. Let us refer to these safety bounds and limits as \textit{constraints}. 
Recent advances in the development of deep RL provide means to allow prior domain knowledge to be encoded in the training processes of neural networks. However, previous efforts on encoding the constraints requires direct modification of the optimization problem. Specifically, the constraints are encoded in the loss function, which may require heavy domain-specific engineering \citep{garcia2015comprehensive}. Consequently, these approaches are not suitable in safety-critical domains where the constraints must be satisfied during learning and thus sample efficiency is an important factor. Although combining expert constraints with neural networks tends to help learning, generating expert constraints remains challenging, understudied and domain-dependent. Further related work is discussed in Appendix~\ref{related_work}. 

In this work we take inspiration from \citep{Fischer2019, Manhaeve, Xu2018} and express our constrains  in first-order logic, which allows for efficient encoding of expert domain knowledge. 
Deep reinforcement learning relies on extensive exploration to generate data, which is highly undesirable when dealing with safety-critical domains. On the other hand, exploration is needed in order to learn and generalize better. 

In this paper, we address both issues. Firstly, we provide a novel way of incorporating constraints in the training processes of deep reinforcement learning. We do not manipulate the current deep learning formulations, i.e. we do not add any extra regularization parameter in the loss function, nor do we rely on any domain-specific engineering. Our approach evaluates the likelihood of constraints being satisfied at each point in time given a state prediction, and this affects the agent’s reward function. Namely, actions that highly satisfy the constraints are encouraged, and actions that do not fully satisfy the constraints are discouraged. Secondly, we use model-based reinforcement learning techniques, which are data-efficient comparing to model-free counterparts. In general, the proposed approach can be seen as analogous to combining system I and system II of the brain, as discussed in Kahneman's \textit{Thinking Fast and Slow} \citep{kahneman2011thinking}. Whilst system I is fast, automatic and intuitive, system II is slower, analytical, and has reasoning capabilities. Hence, we call this approach "System III" as we represent logical constraints and combine them with the high-performing fast deep learning algorithms. 

To show the effectiveness of System III, we conduct experiments on classic control tasks such as the Cart-Pole setup from OpenAI's Gym  \citep{brockman2016openai}, which is a common task in many reinforcement learning algorithms. We show that even a simple constraint on the Cart-Pole system leads to safer exploration and faster convergence. We further conduct experiments on the OpenAI Safety-Gym environments \citep{safetygym}. We show the approach's superiority in safe exploration for a wide arrangement of constraints, hazards and environment configuration in a dynamic setting where the constraints differ across experiments. Further details are discussed in Section \ref{exp}. The contribution of this paper is a novel framework for integrating domain knowledge in the training process of deep RL. Our framework is applicable to any off-the-shelf reinforcement learning algorithm and can be used on top of them to encode domain knowledge and boost sample efficiency significantly while satisfying the constraints. 
\label{intro}

\section{Related work}\label{related_work}

In this section, we compare supervised, unsupervised and RL related works that use human knowledge in the form of constraints.
The integration of constraints as an additional regularizer in the training process of neural networks has achieved considerable attention in recent years, with most work still imposing constraints on the  network's output \citep{Xu2018, Fischer2019}. The main contribution of "Semantics loss" \citep{Xu2018} is the addition of a semantic loss function to the standard neural network loss (e.g. another regularizer), and the design is such that it is equivalent to evaluating some Boolean constraint formula using Weighted Model Counting (WMC) which counts the weights of the solutions to a propositional logic formula \citep{ChaviraD08}. Similar to \citep{Xu2018}, \citep{Fischer2019} defines a non-negative loss function using fuzzy logic to incorporate logical constraints. This loss measures how far the output of the network is from the nearest satisfying solution. Our approach differs from both of these. We compile constraints and add them to the training process of an RL agent; our constraints are motivated by real-world physics, which are crucial for safety-critical domains. Unlike \citep{Xu2018} we do not rely on WMC to evaluate the constraints as they rely on SAT solvers, instead we designed a more suitable metric for constraint evaluation in sequential decision-making tasks. 

One of the closest works to ours is \citep{kogun} comprised of a two-system, a fuzzy rule controller that takes the represented human knowledge constraints and returns a preferred action and a refined module that tunes the suboptimal knowledge. The constraints are represented in fuzzy logic and allow for imprecise policy selection. During the constraint generation, they assume to have perfect knowledge of the state-space, and the constraints fully capture all aspects of the state-space, which is a strong assumption to hold. Our work differs from this: we do not assume to know the full-state space dynamics and do not use constraints to warm-start the policy; further, our constraints are much more general and expressive. To illustrate this generality, consider the rule from \citep{kogun} \textit{Rule: IF $S_1$ is $Ml_1$ and $S_2$ is $Ml_2$ and... and $S_k$ is $Ml_k$ THEN Action is $a_j$}: where $S_i$ are variables that describe different parts of the state, $Ml_i$ is the fuzzy rules corresponding to each $S_i$ and $a_j$ is the action taken. This rule would only be applicable with in fully observable scenarios: an assumption which is unrealistic in many real-world applications. Their approach is no different to hard encoding actions. In our framework, constraints are high level, general, and provide expressiveness. For example, consider when we want to specify that the agent's distance to an undesired object (e.g. traffic light) should be greater than the lower bound or if the distance is less than, the lower bound, the agent should decrease its speed $\varphi(s) = (lb \leq ||s-obj||) \lor (velocity \leq v_{limit} \land ||s-obj|| \leq ub)$. Similarly, there exist other approaches \citep{hasanbeig2018logically,hasanbeig2018logicallynfq,hasanbeig2019reinforcement,hasanbeig2019certified,deepsynth,hasanbeig2020cautious,hasanbeig2020deep,modular_rl_iros,hasanbeig2022certified}
that define safety by the satisfaction of temporal logical formulae of the learnt policy.

Other works in this space \citep{constdqn} propose constrained Q-learning to restrict the action space directly in the Q-update to learn the optimal Q-function; they claim this approach can lead to optimal safe policy in the induced MDP. Our method also differs from this approach, and we do not change anything in the existing deep RL toolbox. However, we believe that some actions become prohibited in our constraint evaluation phase due to low log probability.

\section{Preliminaries}
In this section, we introduce RL formalism and model-based RL methods that we will build upon in the development  of System III. 
We model our problem as \textit{Markov Decision Process} (MDP), which is defined as a tuple  $\langle S, A, P, R, \gamma \rangle$. In the MDP model $S\subseteq\mathbb{R}^{n}$ is a continuous state space, $A\subseteq\mathbb{R}^{m}$ is a continuous action space, and $P:\mathfrak{B}(\mathbb{R}^{n})\times A\times S \rightarrow [0,1]$ is a Borel-measurable conditional transition kernel such that $P(\cdot|s,a)$ is a probability measure of $s\in S$ and $a\in A$ over the Borel space $(\mathbb{R}^{n},\mathfrak{B}(\mathbb{R}^{n}))$, where $\mathfrak{B}(\mathbb{R}^{n})$ is the set of all Borel sets on $\mathbb{R}^{n}$. The transition probability $P$ captures the motion uncertainties of the agent, and it is assumed that $P$ is not known \textit{a priori}. A reward function $R: S \times A \times S \rightarrow \mathbb{R}$ defines a scalar feedback that the agent receives. At each time step $t$ in the environment, the agent observes a state $s_t \in S$, executes an action $a_t \in A$ and transitions to the next state $s_{t+1} \sim P(\cdot | s_t, a_t)$ and receives the reward associated with that action $r_t = R(s_t, a_t, s_{t+1})$. The discount factor $\gamma \in [0, 1]$ is used to weigh the current value of future returns.  A policy $\pi$ is a mapping from the state space to a distribution in $\mathcal{P}(\mathcal{A})$, where $\mathcal{P}(\mathcal{A})$ is the set of probability distributions on subsets of $\mathcal{A}$. A policy is stationary if $\pi(\cdot|s)\in\mathcal{P}(\mathcal{A})$ does not change over time and it is called a deterministic policy if $\pi(\cdot|s)$ is a degenerate distribution. The objective in RL is to find a 
policy $\pi^*$ that maximises the expected discounted sum of rewards,
\begin{equation}\label{utility}
    \mathbb{E_{\pi^*}} [\sum_{t=0}^\infty \gamma^t R(s_t, a_t, s_{t+1})],
\end{equation} where $a_t \sim \pi(\cdot | s_t)$, and $s_{t+1} \sim P(\cdot | s,a)$ \citep{Suttonandbarto98,tdmfmb}. The value of state $s$ under any policy $\pi$, denoted as $V^\pi(s)$, is similarly defined as the expected return starting at state $s$ and following $\pi$ afterwards: 
$
    V^{\pi}(s) = \mathbb{E}_{\pi} \Big[\sum_{t=0}^\infty \gamma^{t}r_{t} \mid s_0 =s \Big].
$
We might drop $\pi$ to simplify notation.
We focus on policy gradient methods, which model and optimise the policy directly. Different policy gradient-based algorithms have been proposed in the literature, e.g., TRPO \citep{TRPO/SchulmanLMJA15} and ACKTR \citep{scaleTRPO:journals/corr/abs-1708-05144}, that learn to update the policy subject to a constraint in the policy space which discourages large differences between successive policies. 
 
 In policy gradient techniques, the key idea is to increase the probability of actions that are associated with higher returns and reduce the probability of actions that lead to lower return until an optimal policy is found. In this paper, we use Asynchronous Advantage Actor Critic policy gradient (A3C) for policy learning \citep{ac3mniha16} combined with Generalized Advantage Estimation (GAE) \citep{GAE}. For a policy $\pi_{\theta}$ where $\theta$ is the neural network parameters. $J(\pi_{\theta})$ is the expected discounted return, and $\nabla_{\theta}J(\pi_{\theta})$ is the gradient of the return with respect to the $\theta$:
$
    \nabla_{\theta}J(\pi_{\theta}) = \mathbb{E_{\pi_{\theta}}} \Big[\sum \nabla_{\theta} \log \pi_{\theta}(s, a)A^{\pi_{\theta}}(s, a)\Big],
    \label{ac}
$
where $A^{\pi_{\theta}}(s, a)$ is the advantage function. The policy gradient algorithm updates the policy network parameter by stochastic gradient ascent 
$\theta \leftarrow \theta + \alpha \nabla_{\theta}J(\pi_{\theta})$, where $\alpha$ is the learning rate.

\section{Constraints Evaluation Using SMT}

To illustrate how human prior knowledge is mapped to first order logic (FL) sentences, consider the example of driving a car, where a learner, i.e. agent, sits in the driver's seat and observes the constraints in the environment, e.g. other cars, designated driving lane, speed limit, humans, traffic signs etc.
Prior to driving, they have prior knowledge about the constraints in the environment, for example, not to crash into humans, traffic signs, and not to get out of the designated lane.
Let us consider the example of not getting out of the designated driving lane and maintaining the speed limit, as shown in Figure~\ref{fig:exp} in Appendix \ref{system3_arc}. 
The prior knowledge for keeping the car in its appropriate lane and respecting the speed limit can be expressed in the form of \textit{($lb \leq$ SPEED LIMIT $\leq ub$) $\land$ ($lb' \leq$ DESIGNATED LANE $\leq ub'$)} defined by a desirable range in Conjunctive Normal Form (CNF) or Disjunctive Normal Form (DNF). 
For instance, if the car speed and the designated lane are in the allowed range, and the distance by which any action moves the car is smaller then the allowed range, any meditate should satisfy the constraint because the agent will still be in the desired range. 

\section{System III Architecture}\label{system3_arc}
\begin{figure}[h!]
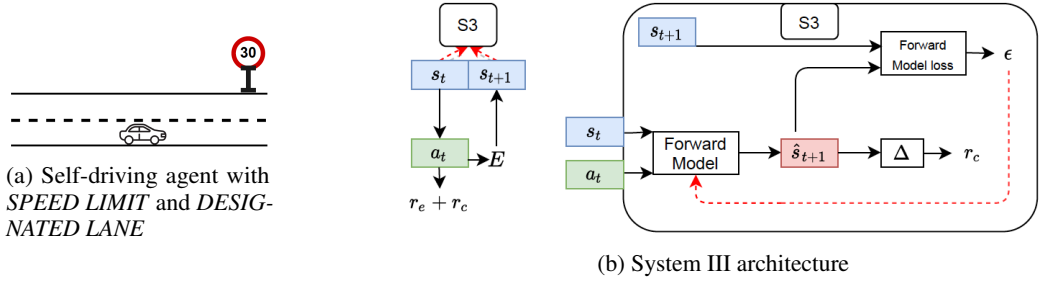

    \centering 
\begin{subfigure}{0.25\textwidth}
  \includegraphics[width=\linewidth]{images/sh.pdf}
  \caption{Self-driving agent with \textit{SPEED LIMIT} and \textit{DESIGNATED LANE}}
  \label{fig:exp}
\end{subfigure}\hspace{5em} 
\begin{subfigure}{0.6\textwidth}
  \includegraphics[width=\linewidth]{images/S3.pdf}
  \caption{System III architecture}
  \label{fig:arc}
\end{subfigure}\hfil 
\caption{S3 short for System III: Figure~\ref{fig:exp} shows a self-driving agent that has to respect the speed limit and designated driving lane. The left-hand side of Figure~\ref{fig:arc} shows the data generation (model free) while the right-hand side of Figure~\ref{fig:arc} shows that the Forward Model takes $s_t$ and $a_t$ as input and returns $\hat{s}_{t+1}$ as output. The Forward Model loss calculates the discrepancy between $s_{t+1}$ and $\hat{s}_{t+1}$ and the constraint satisfaction $\Delta$ calculates the reward $r^c_t$. The black arrows indicate forward pass while, the red arrow indicates backward pass.}
\end{figure}

\section{Running Example}\label{running_example}
Slippery grid world example below:

\begin{figure}[!ht]
	\centering
	\scalebox{0.8}[1]{
	\hspace{-5.5mm}\subfloat[]{{\includegraphics[width=0.4\linewidth]{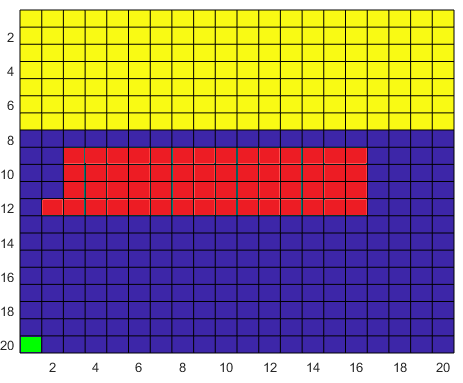}}}}\\
	\subfloat[]{{\includegraphics[width=0.40\linewidth]{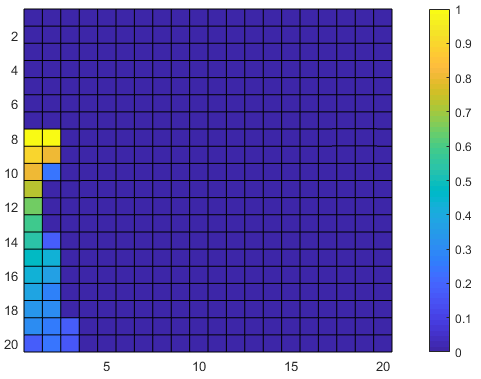}}}%
	\subfloat[]{{\includegraphics[width=0.40\linewidth]{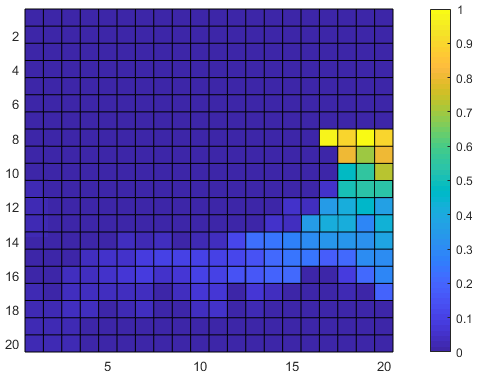}}}%
	\caption{(a) Slippery grid world example, where yellow is the $\mathit{target}$, red is $\mathit{unsafe}$, blue is $\mathit{safe}$, and green is the initial state $s_0$; Safety and performance tuning with FL: (b)~value function $V(s)$ with $lb=0$; and (c)~value function with $lb>0$ where the FL formula is $\varphi(s) = lb \leq ||s-\mathit{unsafe}||$, and $||s-\mathit{unsafe}||$ is the agent's distance to unsafe (red) states.}%
	\label{fig_bridge}
\end{figure}
\section{Combining Logical Constraints and Reinforcement Learning}

\textbf{High-Level FL Constraint:}
We take motivation from real-world settings, where expressive yet intuitive constraints are required to fully capture the desired behavior while are easily understandable for humans. For example, consider the task of parking between two objects. Formally, given the agent state $s$, the FL constraint is: $$\varphi(s) = \forall~ \mathit{object_i}:~ lb' \leq ||s-\mathit{object_i}||,$$ where 
$||s-\mathit{object_i}||$ is the agent's distance to the objects while $lb'$ is the lower bound on the distance. 

{The Running Example} is shown in Figure~\ref{fig_bridge} in Appendix~\ref{running_example} and is a moving robot in a grid. Let the grid be a $ 20 \times 20 $ square over which the robot moves. In this setup, the robot location is the MDP state $s \in {S} $. At each state $s \in {S}$ the robot has a set of actions $ {A}=\{\mathit{left},\mathit{right},\mathit{up},\mathit{down},\mathit{stay}\}$ using which the robot is able to move to other states (e.g. $s'$) with the probability of $P(s,a,s'), a \in \mathcal{A}$. At each state $s \in {S}$, the actions available to the robot are either to move to a neighbour state $s' \in {S}$ or to stay at the state $s$. In this example, we assume for each action the robot chooses, there is a probability of $85\%$ that the action takes the robot to the correct state and $15\%$  that the action takes the robot to a random state in its neighbourhood, including its current state. The property of interest in this example is an FL formula: $\varphi(s) = \forall~ \mathit{unsafe}_i:~ ||s-\mathit{unsafe}_i|| \leq lb$, where $||s-\mathit{unsafe}_i||$ is the agent's distance to any unsafe (red) state $\mathit{unsafe}_i$.

\textbf{System III architecture:}
System III comprises two reciprocating subsystems: a system that learns to model the next state $s_{t+1}$ and a system that evaluates the constraints on $s_{t+1}$. The ability of the latter subsystem to evaluate the constraints depends on the former subsystem's ability to accurately model the next state. We further define the total reward $r_t$ to be the sum of reward returned by the environment ($r^e_t$) and reward returned by the degree of constraints satisfaction ($r^c_t$) resulting in:
    $r_t = {r^e_t} + {r^c_t}$. 
%

We use a policy $\pi(\cdot|s_t;\theta_{p})$ representing an actor neural network, where $s_t$ is the current state, and $\theta_{p}$ refers to the weights of the network. An agent at state $s_t$ takes the action $a_t \sim \pi(\cdot|s_t;\theta_{p})$ sampled from the policy. The parameters of the policy network $\theta_p$ are optimised to maximize the sum of discounted expected return in (\ref{utility}). 
In order to ensure that the agent explores the state space sufficiently to learn and satisfy the necessary constraints, we construct a model to take as input the state $s_t$ and action $a_t$ and returns the distribution of next state at $t+1$. This is also known as forward dynamics. Formally: ${\hat{s}_{t+1}} \sim f(\cdot | s_t, a_t; \theta_{F})$, where $\hat{s}_{t+1}$ is the estimate of ${s}_{t+1}$ and $\theta_F$ is the forward model (e.g. a neural network) parameters. From the forward dynamics we optimize the parameter $\theta_{F}$ by minimizing the mean squared loss function:
\begin{equation}\label{lf}
    L_f=\sum_{t}^{}(f(s_t, a_t) - (s_{t+1}))^2.
\end{equation}
With the model $f$ trained via mean squared loss function , we can evaluate the constraints satisfaction at each time step. Consider the running example FL constraint $\varphi(s) = lb \leq ||s-\mathit{unsafe}||$. Define $\Delta: S \times A \times S \rightarrow \mathbb{R}$ as the constraint evaluation function. At each time step $t$, the agent observes a state $s_t \in S$, chooses an action $a_t \in A$ and the forward model $f$ outputs ${\hat{s}_{t+1}} \sim f(\cdot | s_t, a_t; \theta_{F})$. The constraint reward is then defined as:
\begin{center}
  ${r^c_t} = \Delta(s_t,a_t,\hat{s}_{t+1})=
\begin{cases}
1 ~~~~\text{if}~~ \varphi(\hat{s}_{t+1})\\
0 ~~~~\text{otherwise}
\end{cases}$. 
\end{center}



We can write the overall optimization problem as a combination of (\ref{utility}) and (\ref{lf}):
\begin{equation}\label{combined_loss}
    \min_{\theta_{p}\theta_{F}} [-\lambda \mathop{\mathbb{E}_{\pi_{(s_t; \theta_p)}}}[\sum_t \gamma r_t] + \beta L_F ],
\end{equation}
where $0 \leq \beta \leq 1$ is a scalar that weighs the forward model loss  and $\lambda > 0$ is a scalar that weighs the importance of the policy gradient loss against the importance of satisfying the constraint.

\section{Experiments setup and Results}
\label{exp}


In the running example, in order to get to the target state the agent has to cross a bridge (Fig.~\ref{fig_bridge}a) surrounded by unsafe states. The grid is slippery, namely from the agent's perspective, when it takes an action it usually moves to the intended cell, but there is an \textit{unknown} probability that the agent is moved to a random neighbour cell. However, the trained model $\hat{P}$ initially advises the agent that it can always move to the correct state and this is the dynamics known to the agent. The initial state of the agent is bottom left.
For the simulated physical environments we consider OpenAI Gym \citep{gym} and Safety Gym \citep{safetygym}.
OpenAI's Gym environment comprises a set of toolkits for developing and comparing reinforcement learning algorithms. It contains tasks ranging from control to Attari. For this paper, we focus on the continuous control task CartPole \citep{Suttonandbarto98}. 
In Safety Gym, we run experiments using the Point, Car and Doggo robot \citep{safetygym} while varying the number of constraints in the environment with constant goal task. In both environments, the agent interacts with the environment and is rewarded based on the degree to which it satisfies the constraints. We leave the case for sparse reward setting for future work. Consider Figure~\ref{fig:7}, where the agent task is to press the highlighted button while avoiding hazards. We represent such scenario via a general FL formula:
$\varphi(s) = \{(\forall~{\mathit{hazard}_i}: lb_h \leq ||s-\mathit{hazard}_i|| \leq  max\_distance\_pair\_button) \land \\
 (\forall~{\mathit{goal\_button}_i}: ||s-\mathit{goal\_button}_i|| \leq  max\_distance\_pair\_button)\}.$

\begin{figure}[h!] \centering \subfloat[\centering Performers on CartPole with no Constraints]{{\includegraphics[width=5cm]{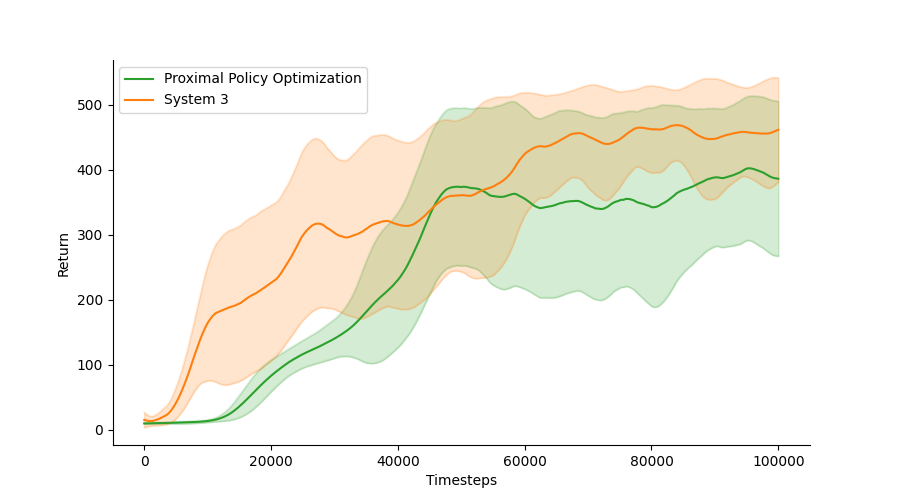} }}%
\qquad \subfloat[\centering CartPole Constraint Violation]{{\includegraphics[width=5cm]{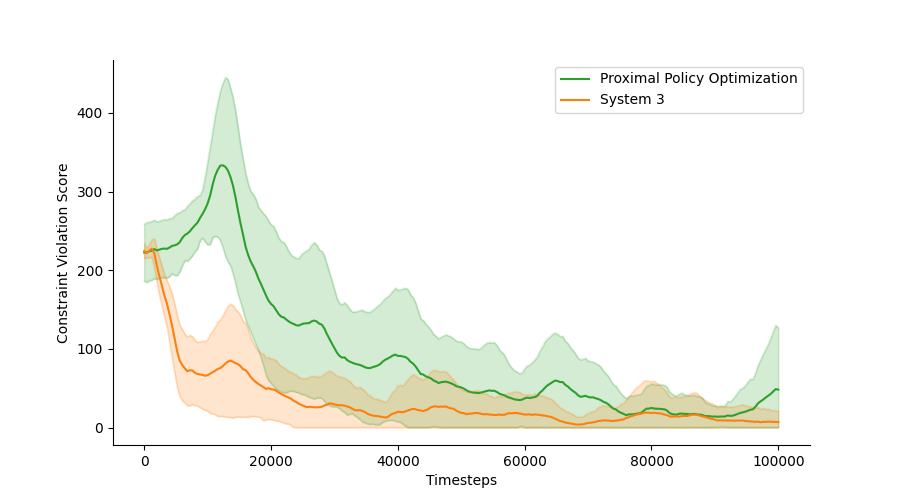} }}%
\qquad \subfloat[\centering Performance of System III vs PPO with no constraints]{{\includegraphics[width=5cm]{images/Cartpole-Scores-S3NoConstraint.png} }}%
\qquad \subfloat[\centering Constraint violations of System III vs PPO with no constraints]{{\includegraphics[width=5cm]{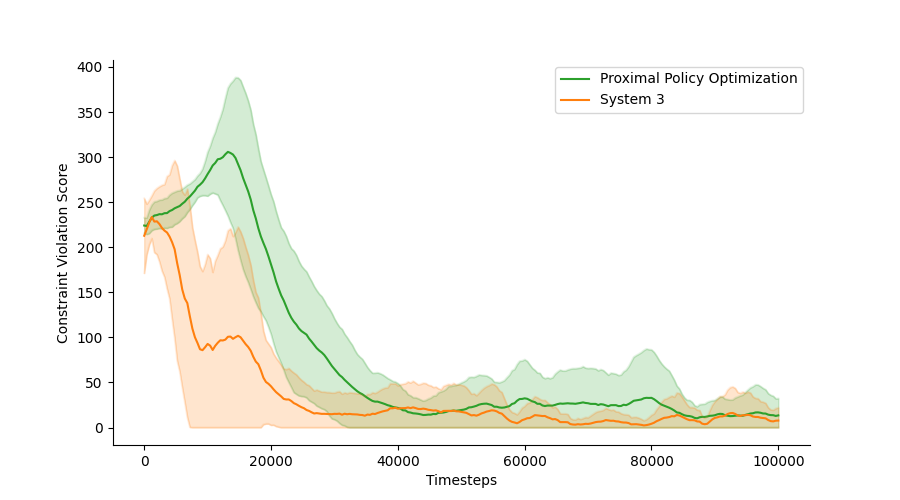} }}%
\caption{\texttt{cartpole-v1} trained for 1e6 time steps (max score is 500). Each training loop has a length of 100 time steps and batch size of 20, hidden layers each has 64 nodes. Every 400 time steps, the model is saved and evaluated for 1000 time steps in “test runs”, i.e., no training. The score for each of the episodes that are completed are added in the test runs. The constraints are added up across all of the time steps in the test runs. The score plots are averaged between 10 different instances of the model at each time step, and the graph is smoothed with a 20 step rolling mean. The error plots are the same but with a 10 step rolling mean. The constraint are placed on the state space, e.g., angle and position.} \label{fig:cartpole} 
\end{figure}

Constraints are defined as an 'allowable' subspace of the state space.  In the CartPole experiment we deliver $d$-step accumulated reward every $d$ time steps. The constraints on the in CartPole is: $\varphi(s) =(lb \leq x \leq ub) \land (lb \leq k \leq ub)$ where $x$ corresponds to the cart position, and $k$ corresponds to the pole angle at tip. We evaluate our algorithm under different delayed steps. Figure \ref{fig:cartpole} plots the results under a sparse reward setting. Similarly, in Safety-Gym, we provide $r^c_t$ after each immediate action and $r^e_t$ is fully ignored as $r^c_t$ captures the degree of satisfaction in each state. Table \ref{tab:result_all_env} shows the average episodic mean return along with constraint violation (i.e. 1 - constraint satisfaction), where $0$ corresponds to $0\%$ constraint satisfaction and $1$ corresponds to $100\%$ satisfaction at evaluation. We train our system's combined objective in \ref{combined_loss} with $\lambda$ = 0.15, $\beta$ = 0.3 and learning rate of $1e^{-3}$. We observe there to exist an inverse relationship between  the reward and constraint satisfaction. Unconstrained algorithms (e.g. PPO and TRPO) achieves higher return at the cost of high degree of constraint violation. However, PPO and TRPO's Lagrangian counter part which follows the adaptive penalty to enforce constraints achieves higher degree of constraint satisfaction. We observe a an inverse relationship between achieving higher reward and acting safely. System 3 is able to act safely after a small amount of interaction with the environment at a cost of slightly lower return compared to the other methods which achieves slightly higher return but does significantly worse at satisfying the constraints. We hypothesis this is due to the nature of the the environment reward function not being able to capture the actual desire of the designer and what it intends it to do. We show that our method achieves high constraint satisfaction (95\%) as shown in \ref{tab:result_all_env} in Appendix~\ref{resultsappendix}, compared to popular baselines designed to deal with constrained MDPs.

\label{resultsappendix}

\begin{table}[!ht]
    \centering
\begin{tabular}{cccc} \toprule \centering
    {Method} & {Mean return} & Constraint Satisfaction\\ \midrule
    PPO  &  1.0  & 0.16\\
    PPO-Lagrangian  &  0.15   & 0.83\\
    TRPO  &   1.0 & 0.42 \\
    TRPO-Lagrangian   & 0.61    & 0.83\\ \midrule
   
   System 3   & 0.75 & \pmb{0.95}\\ \bottomrule
\end{tabular}
\caption{Normalized metrics from training averaged all environments and three random seeds per environment.}
    \label{tab:result_all_env}
\end{table}
\begin{table}
\centering
\begin{tabular}{clllll} \toprule
\multicolumn{2}{l}{Constraint Type} & \multicolumn{2}{c}{Static}       & \multicolumn{2}{c}{Static \& Moving} \\ \midrule
\multicolumn{1}{l}{Method}  & Agent & Return & Constraint Sat & Return   & Constraint Sat   \\
\midrule
\multirow{3}{*}{PPO}        & Point & 1.0       & 0.16                         & 1.0         & 0.11                         \\
                            & Car   & 1.0       &  0.26                       & 1.0       &0.23                           \\
                            & Doggo & 1.0       & 0.08                        & 1.0         & 0.17                       \\ \midrule
\multirow{3}{*}{PPO-Lagrangian}  & Point & 0.36        &0.67  & 0.21         &0.62                           \\
                            & Car   & 0.45       & 0.73                        & 0.55         & 0.64                        \\
                            & Doggo &0.71       &  0.59                       &  0.88        &  0.62                      \\ \midrule 
\multirow{3}{*}{TRPO}        & Point &  1.0      & 0.32  & 0.99         & 0.28                          \\
                            & Car   &  1.0     &  0.41                       &  1.0        &  0.45                         \\
                            & Doggo &  1.0      & 0.64                        & 1.0         & 0.43                       \\ \midrule 
\multirow{3}{*}{TRPO-Lagrangian}        & Point & 0.41       &0.73 &0.56       &0.84                           \\
                            & Car   &  0.62      &  0.88                       &  0.73        &  0.88                         \\
                            & Doggo &  0.69      &    0.89                     & 0.67         &  0.84                       \\ \midrule
\multirow{3}{*}{System3}        & Point &0.73        &\textbf{ 0.96}& 0.68       &\textbf{0.94}                          \\
                            & Car   & 0.72     & \textbf{0.96}                        & 0.75       &\textbf{0.95}                         \\
                            & Doggo &  0.81      &\textbf{0.94} &  0.82        &\textbf{0.93}                       \\
                            \bottomrule
\end{tabular}
\caption{Agents: Mean return (return) and constraint satisfaction. Normalized metrics from training averaged all environments and three random seeds per environment}
\end{table}

\section{Conclusions And Future Work}
In this paper we propose a novel framework for incorporating constraints in the training processes of deep reinforcement learning.  Our approach evaluates the likelihood of constraints being satisfied at each point in time given a state prediction, and this affects the agent’s reward function. The future work should consider changing the environments and keeping the constraints constant and learn the constraints directly from the environment. From a novel safety and alignment perspective this work provides a solution for outer alignment, however future research would need to address issues that might arise with inner alignment. 

\clearpage
\bibliography{iclr2022_conference}
\bibliographystyle{plain}
\newpage
\appendix
\section*{Appendix}

\section{OpenAI SafetyGym}\label{safety_gym}
\begin{figure}[!ht]
    \centering 
\begin{subfigure}{0.3\textwidth}
  \includegraphics[width=\linewidth]{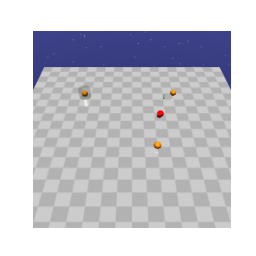}
  \caption{Point Agent with press task}
  \label{fig:7}
\end{subfigure}\hfil 
\begin{subfigure}{0.3\textwidth}
  \includegraphics[width=\linewidth]{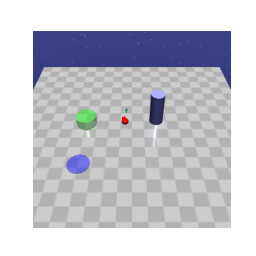}
  \caption{Point Agent with position task}
  \label{fig:8}
\end{subfigure}\hfil 
\begin{subfigure}{0.3\textwidth}
  \includegraphics[width=\linewidth]{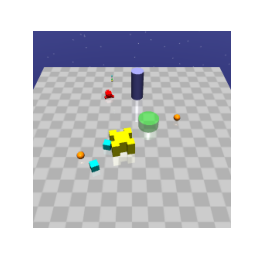}
  \caption{Point Agent with push task}
  \label{fig:9}
\end{subfigure}
\medskip
\begin{subfigure}{0.3\textwidth}
  \includegraphics[width=\linewidth]{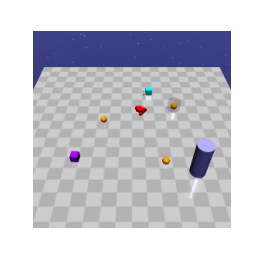}
  \caption{Car Agent with press task}
  \label{fig:1}
\end{subfigure}\hfil 
\begin{subfigure}{0.3\textwidth}
  \includegraphics[width=\linewidth]{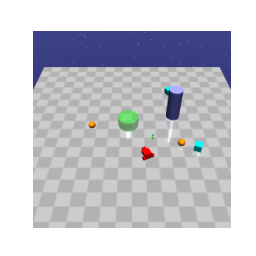}
  \caption{Car Agent with position task}
  \label{fig:2}
\end{subfigure}\hfil 
\begin{subfigure}{0.3\textwidth}
  \includegraphics[width=\linewidth]{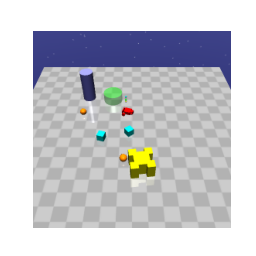}
  \caption{Car Agent with push task}
  \label{fig:3}
\end{subfigure}
\medskip
\begin{subfigure}{0.3\textwidth}
  \includegraphics[width=\linewidth]{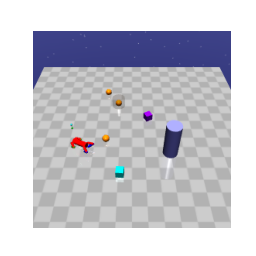}
  \caption{Quadruped Agent with press task}
  \label{fig:4}
\end{subfigure}\hfil 
\begin{subfigure}{0.3\textwidth}
  \includegraphics[width=\linewidth]{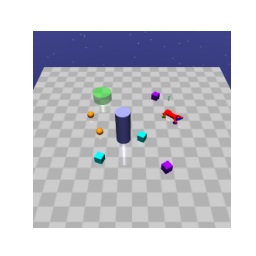}
  \caption{Quadruped Agent with position task}
  \label{fig:5}
\end{subfigure}\hfil 
\begin{subfigure}{0.3\textwidth}
  \includegraphics[width=\linewidth]{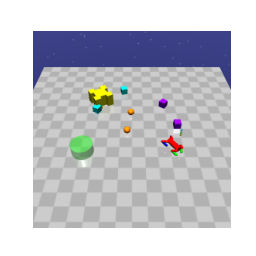}
  \caption{Quadruped Agent with push task}
  \label{fig:6}
\end{subfigure}
\caption{Environment Agent and Constraint Configuration:
All constraint elements represent scenarios for the agent to avoid; they pose different challenges for the agent by virtue of having different dynamics. To illustrate the contrast: hazards (purple circle) provide no physical obstacle, vases (blue box) are movable obstacles, pillars (tall purple box) are immovable obstacles, buttons (orange button) can sometimes be perceived as goals, and gremlins (purple box) are actively-moving obstacles.}
\label{fig:images}
\end{figure}
\clearpage

\end{document}